# Nonparametric Bayesian Factor Analysis for Dynamic Count Matrices

**Ayan Acharya**
Dept. of ECE, UT Austin

**Joydeep Ghosh**
Dept. of ECE, UT Austin

**Mingyuan Zhou**
Dept. of IROM, UT Austin

## Abstract

A gamma process dynamic Poisson factor analysis model is proposed to factorize a dynamic count matrix, whose columns are sequentially observed count vectors. The model builds a novel Markov chain that sends the latent gamma random variables at time $(t-1)$ as the shape parameters of those at time $t$, which are linked to observed or latent counts under the Poisson likelihood. The significant challenge of inferring the gamma shape parameters is fully addressed, using unique data augmentation and marginalization techniques for the negative binomial distribution. The same nonparametric Bayesian model also applies to the factorization of a dynamic binary matrix, *via* a Bernoulli-Poisson link that connects a binary observation to a latent count, with closed-form conditional posteriors for the latent counts and efficient computation for sparse observations. We apply the model to text and music analysis, with state-of-the-art results.

## 1 INTRODUCTION

There has been growing interest in analyzing dynamic count and binary matrices, whose columns are data vectors that are sequentially collected over time. Such data appear in many real world applications, such as text analysis, social network modeling, audio and language processing, and recommendation systems. Count data are discrete and nonnegative, have limited ranges, and often present overdispersion; binary data only have two possible values: 0 and 1; and both kinds of data commonly appear in big matrices that are extremely sparse. While the classical matrix factorization method using the Frobenius norm is effective for factorizing real matrices [1, 2, 3, 4, 5, 6, 7], its inherent Gaussian assumption is often overly restrictive for modeling count and binary matrices. Exploiting well-developed techniques for Gaussian data, one usually considers connecting a count observation to a latent Gaussian random variable using the lognormal-Poisson link, and connecting a binary observation using the probit or logit links. These generalized linear model [8] based approaches, however, might involve heavy computation and lack intuitive interpretation of the inferred factorization.

Despite these disadvantages, latent Gaussian based approaches are commonly used to analyze count and binary data. This is particularly true for dynamic modeling, since inference techniques for linear dynamical systems such as the Kalman filter are well developed, and can be readily applied once the dynamic count/binary data are transformed into the latent Gaussian space. For example, to analyze the temporal evolution of topics in a corpus, the dynamic topic model draws the topic proportion at each time stamp from a logistic normal distribution, whose parameters are chained in a state space model that evolves with Gaussian noise [9]. Although the dynamic topic model is a discrete latent variable model, to model the topic proportion that explains the number of words assigned to a topic in a document, which is a count, it chooses to use the logistic normal link and imposes a temporal smoothness in the latent Gaussian space.

Rather than modeling the temporal evolution of count and binary data in the latent Gaussian space using a linear dynamical system, we consider a fundamentally different approach: we directly chain the positive Poisson rates of the count or binary data in a state space model that evolves with gamma noise. More specifically, we build a gamma Markov chain that sends $\theta_{t-1}$, a latent gamma random variable at time $t-1$, as the shape parameter of the latent gamma random variable at time $t$ as $\theta_t | \theta_{t-1} \sim \text{Gam}(\theta_{t-1}, 1/c)$; at each time point, we use $\theta_t$ as the Poisson rate for a count as $n_t \sim \text{Pois}(\theta_t)$; and the counts $\{n_t\}_t$ are conditionally independent given $\{\theta_t\}_t$. If the observation is binary, then we assume the Bernoulli random variable is generated by thresholding a latent count as $b_t = \mathbf{1}(n_t \geq 1)$, which means $b_t = 1$ if $n_t \geq 1$ and $b_t = 0$ if $n_t = 0$. We call this count-to-binary link function as the Bernoulli-Poisson link, under which the conditional posterior of the latent count follows a truncated Poisson distribution.

To apply the gamma Markov chain to dynamic count and binary matrix factorization, we extend it to a multivariate setting, which is integrated into a discrete latent variable model called Poisson factor analysis [10]. Specifically, we factorize the observed dynamic count (binary) matrix under the Poisson (Bernoulli-Poisson) likelihood, and chain the latent factor scores across time, where a gamma distributed factor score is linked via a Poisson distribution to a *latent* count that counts how many times the corresponding factor is used by the corresponding observation. To avoid tuning the latent dimension of factorization, we also employ a gamma process to automatically infer the number of factors, which can be potentially infinite. The key challenge for this unconventional Markov chain is to infer the gamma shape parameters, for which we discover a simple and effective solution.

The paper makes the following contributions: 1) We construct a novel gamma Markov chain to model dynamic count and binary data. 2) We provide closed-form update equations to infer the parameters of the gamma Markov chain, using novel data augmentation and marginalizing techniques. 3) We integrate the gamma Markov chain into Poisson factor analysis to analyze dynamic count matrices. 4) We factorize a dynamic binary matrix under the Bernoulli-Poisson likelihood, with extremely efficient computation for sparse observations. 5) We apply the developed techniques to real world dynamic count and binary matrices, with state-of-the-art results.

## 2 PRELIMINARIES

**Negative Binomial Distribution:** The negative binomial (NB) distribution $m \sim \text{NB}(r, p)$, with probability mass function (PMF) $\Pr(M = m) = \frac{\Gamma(m+r)}{m!\Gamma(r)}p^m(1-p)^r$, where $m \in \mathbb{Z}$ and $\mathbb{Z} = \{0, 1, \ldots\}$, can be augmented into a gamma-Poisson construction as $m \sim \text{Pois}(\lambda)$, $\lambda \sim \text{Gam}(r, p/(1-p))$, where the gamma distribution is parameterized by its shape $r$ and scale $p/(1-p)$. It can also be augmented under a compound Poisson representation as $m = \sum_{t=1}^{l} u_t$, $u_t \stackrel{iid}{\sim} \text{Log}(p)$, $l \sim \text{Pois}(-r\ln(1-p))$, where $u \sim \text{Log}(p)$ is the logarithmic distribution [11].

**Lemma 2.1** ([12]). *If $m \sim \text{NB}(r, p)$ is represented under its compound Poisson representation, then the conditional posterior of $l$ given $m$ and $r$ has PMF:*

$$\Pr(l = j | m, r) = \frac{\Gamma(r)}{\Gamma(m+r)}|s(m, j)|r^j, \; j = 0, 1, \ldots, m,$$

*where $|s(m, j)|$ are unsigned Stirling numbers of the first kind. We denote this conditional posterior as $l \sim \text{CRT}(m, r)$, a Chinese restaurant table (CRT) count random variable, which can be generated via $l = \sum_{n=1}^{m} z_n, z_n \sim \text{Bernoulli}(r/(n-1+r))$.*

**Gamma Process:** The gamma Process [13, 14] $G \sim \text{GaP}(c, G_0)$ is a completely random measure [15, 16] defined on the product space $\mathbb{R}_+ \times \Omega$, where $\mathbb{R}_+ = \{x : x > 0\}$, with concentration parameter $c$ and a finite and continuous base measure $G_0$ over a complete separable metric space $\Omega$, such that $G(A_i) \sim \text{Gam}(G_0(A_i), 1/c)$ are independent gamma random variables for disjoint partition $\{A_i\}_i$ of $\Omega$. The Lévy measure of the gamma process can be expressed as $\nu(drd\omega) = r^{-1}e^{-cr}drG_0(d\omega)$. Since the Poisson intensity $\nu^+ = \nu(\mathbb{R}_+ \times \Omega) = \infty$ and $\int_{\mathbb{R}_+ \times \Omega} r\nu(drd\omega)$ is finite, following [14], a draw from the gamma process consists of countably infinite atoms, which can be expressed as $G = \sum_{k=1}^{\infty} r_k \delta_{\omega_k}$, where $(r_k, \omega_k) \stackrel{iid}{\sim} \pi(drd\omega)$ and $\pi(drd\omega)\nu^+ \equiv \nu(drd\omega)$.

**Poisson Factor Analysis:** A large number of discrete latent variable models can be united under Poisson factor analysis (PFA) [10], which factorizes a count matrix $\mathbf{N} \in \mathbb{Z}^{V \times T}$ under the Poisson likelihood as $\mathbf{N} \sim \text{Pois}(\mathbf{\Phi\Theta})$, where $\mathbf{\Phi} \in \mathbb{R}_+^{V \times K}$ is the factor loading matrix or dictionary and $\mathbf{\Theta} \in \mathbb{R}_+^{K \times T}$ is the factor score matrix. A wide variety of algorithms, although constructed with different motivations and for distinct problems, can all be viewed as PFA with different prior distributions imposed on $\mathbf{\Phi}$ and $\mathbf{\Theta}$. For example, non-negative matrix factorization [17, 18], with the objective to minimize the Kullback-Leibler divergence between $\mathbf{N}$ and its factorization $\mathbf{\Phi\Theta}$, is essentially PFA solved with maximum likelihood estimation. Latent Dirichlet allocation [19] is equivalent to PFA, in terms of both block Gibbs sampling and variational inference, if Dirichlet distribution priors are imposed on both $\phi_k \in \mathbb{R}_+^V$, the columns of $\mathbf{\Phi}$, and $\theta_t \in \mathbb{R}_+^K$, the columns of $\mathbf{\Theta}$. The gamma-Poisson model [20, 21, 22] is PFA with gamma priors on $\mathbf{\Phi}$ and $\mathbf{\Theta}$. A family of NB processes, such as the beta-NB [10, 23] and gamma-NB processes [12, 24], impose different gamma priors on $\{\theta_{tk}\}$, the marginalization of which leads to differently parameterized NB distributions; for example, the beta-NB process imposes $\theta_{tk} \sim \text{Gam}(r_t, p_k/(1-p_k))$, where $\{p_k\}_{1,\infty}$ are the weights of the countably infinite atoms of the beta process [25], and the gamma-NB process imposes $\theta_{tk} \sim \text{Gam}(r_k, p_t/(1-p_t))$, where $\{r_k\}_{1,\infty}$ are the weights of the countably infinite atoms of the gamma process. Both the beta-NB and gamma-NB process PFAs are nonparametric Bayesian models that allow $K$ to grow without limit.

**Related Dynamic Models for Count Data Analysis:** The dynamic rank factor model (DRFM [26]) performs analysis of multiple ordinal time series where the latent dynamics are modeled using a transition matrix and the observations are sampled from another latent variable that, in turn, is sampled using a normal distribution centered around such latent space. The dynamic-HDP [27] models the temporal evolution of Dirichlet random probability measures and is designed for mixture models whose mixture weights smoothly evolve over time. In [28], the authors model the temporal evolution of relational data, with the temporal smoothness constraint on latent variables imposed in the Gaussian space. The online multiscale dynamic topic model [29] analyzes the evolution of *sets* of documents. The proposed gamma process dynamic PFA (GP-DPFA) is a factor model that describes the temporal evolution of latent factor scores (not normalized random measures like in the dynamic-HDP); at each time point, it models a single count/binary vector (not a set of documents like in a dynamic topic model). Interestingly, in [30], the author models groups of related count time series through a shared latent Gaussian space, though the results from that paper are not reproducible as neither the code nor the datasets are publicly available. To impose temporal smoothness in the frequency domain for audio processing, [31] considers chaining latent variables across successive time frames via the gamma scale parameters, whereas GP-DPFA chains latent variables via the gamma shape parameters only.

# 3 GAMMA PROCESS DYNAMIC POISSON FACTOR ANALYSIS

Consider a dynamic count matrix $\mathbf{N} \in \mathbb{Z}^{V \times T}$, whose $T$ columns are sequentially observed $V$-dimensional count vectors. We consider a modified version of PFA as $\mathbf{N} \sim \text{Pois}(\boldsymbol{\Phi}\boldsymbol{\Lambda}\boldsymbol{\Theta})$, where $\boldsymbol{\Lambda} = \text{diag}(\boldsymbol{\lambda})$ and $\boldsymbol{\lambda} = (\lambda_1, \ldots, \lambda_\infty)$ is a vector representing the strengths of the countably infinite latent factors. Further, a gamma process $G \sim \text{GaP}(c, G_0)$ is considered, a draw from which is expressed as $G = \sum_{k=1}^\infty \lambda_k \delta_{\boldsymbol{\phi}_k}$, where $\boldsymbol{\phi}_k \in \Omega$ is an atom drawn from a $V$-dimensional base distribution $G_0(d\boldsymbol{\phi}_k)/G_0(\Omega) = \text{Dir}(d\boldsymbol{\phi}_k; \eta, \ldots, \eta)$ and $\lambda_k = G(\boldsymbol{\phi}_k)$ is the associated weight. We mark each atom $\boldsymbol{\phi}_k$ with a constant $\theta_{(-1)k} = 0.01$, and then generate a gamma Markov chain by letting:

$$\theta_{tk}|\theta_{(t-1)k} \sim \text{Gam}(\theta_{(t-1)k}, 1/c_t), \quad t = 0, \ldots, T.$$

We then integrate the weights of the gamma process $\{\lambda_k\}$ and the infinite-dimensional gamma Markov chain into a gamma process dynamic Poisson factor analysis (GP-DPFA) model as:

$$n_{vt} = \sum_{k=1}^\infty n_{vtk}, \ n_{vtk} \sim \text{Pois}(\lambda_k \phi_{vk} \theta_{tk}),$$
$$\boldsymbol{\phi}_k \sim \text{Dir}(\eta_1, \ldots, \eta_V), \quad \theta_{tk} \sim \text{Gam}(\theta_{(t-1)k}, 1/c_t),$$
$$G \sim \text{GaP}(c, G_0), \quad c_t \sim \text{Gam}(e_0, 1/f_0).$$

We further let both the concentration parameter $c$ and mass parameter $\gamma_0 = G_0(\Omega)$ be drawn from $\text{Gam}(e_0, 1/f_0)$. Note that under the regular setting where different columns of $\boldsymbol{\Theta}$ are independently modeled, the parameterization $\mathbf{N} \sim \text{Pois}(\boldsymbol{\Phi}\boldsymbol{\Lambda}\boldsymbol{\Theta})$ is not a strict generalization of the beta-NB process PFA described in [10, 12]: if one follows the beta-NB process to let $\theta_{tk} \sim \text{Gam}(r_t, p_k/(1-p_k))$, and $\lambda_k$ is assumed to be independent from $\theta_{tk}$, then $\tilde{\theta}_{tk} := \lambda_k \theta_{tk} \sim \text{Gam}(r_t, q_k/(1-q_k))$, where $q_k = \frac{\lambda_k p_k}{1+(\lambda_k-1)p_k}$; thus $\boldsymbol{\Lambda}$ are redundant and can be absorbed into $\boldsymbol{\Theta}$ as $\mathbf{N} \sim \text{Pois}(\boldsymbol{\Phi}\widetilde{\boldsymbol{\Theta}})$. Whereas in this paper, with the column index $t$ corresponding to time, for tractable inference, it becomes necessary to use the modified representation to impose a temporal smoothness constraint for consecutive columns, which are no longer assumed to be independent, as discussed below.

## 3.1 Inference *via* Gibbs Sampling

Though GP-DPFA supports a countably infinite number of latent factors, in practice, it is impossible to instantiate all of them. Common approaches for exact inference for a nonparametric Bayesian model involve either marginalizing out the underlying stochastic process [32, 33] or using slice sampling to adaptive truncate the number of atoms [34]. For simplicity, in this paper, we consider a finite approximation of the infinite model by truncating the number of factors to $K$, with $\lambda_k \sim \text{Gam}(\gamma_0/K, 1/c)$, which approaches the original infinite model as $K \to \infty$. Despite the significant challenge presented in inferring the gamma shape parameters, generalizing the data augmentation and marginalization techniques unique to the negative binomial distribution [12, 24], we are able to derive closed-form Gibbs sampling update equations.

Exploiting the property that $\sum_{v=1}^V \phi_{vk} = 1$ for any $k$, the likelihood of the latent counts, conditioned on $(\boldsymbol{\Phi}, \boldsymbol{\Theta}, \boldsymbol{\lambda})$, can be expressed as in Table 1. Below we let $n_{v\cdot k} := \sum_t n_{vtk}$, $n_{\cdot tk} := \sum_v n_{vtk}$, and $n_{\cdot \cdot k} := \sum_v \sum_t n_{vtk}$.

***Sample*** $n_{vtk}$: Using the relationship between the Poisson and multinomial distributions, as in Lemma 4.1 of [10], given the observed counts and latent parameters, we have:

Nonparametric Bayesian Factor Analysis for Dynamic Count Matrices

Table 1: Likelihood of GP-DPFA

$$P(\{(n_{vtk})_{k=1}^K\}|\boldsymbol{\Phi},\boldsymbol{\Theta},\boldsymbol{\Lambda}) = \prod_{t=1}^T\prod_{v=1}^V\prod_{k=1}^K \frac{(\lambda_k\phi_{vk}\theta_{tk})^{n_{vtk}}}{n_{vtk}!}e^{-\lambda_k\phi_{vk}\theta_{tk}}.$$
$$= \left(\prod_{t=1}^T\prod_{v=1}^V\prod_{k=1}^K \frac{1}{n_{vtk}!}\right)\prod_{k=1}^K\left\{\left(\prod_{v=1}^V \phi_{vk}^{n_{v\cdot k}}\right)\left(\prod_{t=1}^T \theta_{tk}^{n_{\cdot tk}}\lambda_k^{n_{\cdot tk}}e^{-\lambda_k\theta_{tk}}\right)\right\}.$$

$$\left((n_{vtk})_{k=1}^K|-\right) \sim \text{Mult}\left(n_{vt},\left(\frac{\lambda_k\phi_{vk}\theta_{tk}}{\sum_k \lambda_k\phi_{vk}\theta_{tk}}\right)_{k=1}^K\right). \quad (1)$$

**Sample $\phi_k$:** Using the Dirichlet-multinomial conjugacy and the likelihood in Table 1, the conditional posterior of $\phi_k$ can be expressed as

$$(\phi_k|-) \sim \text{Dir}(\eta+n_{1\cdot k},\ldots,\eta+n_{V\cdot k}). \quad (2)$$

**Sample $\lambda_k$:** Since in the likelihood $n_{\cdot tk} \sim \text{Pois}(\lambda_k\theta_{tk})$, using the gamma-Poisson conjugacy, the conditional posterior of $\lambda_k$ can be expressed as

$$(\lambda_k|-) \sim \text{Gam}\left(n_{\cdot\cdot k}+\frac{\gamma_0}{K},\frac{1}{c+\sum_t \theta_{tk}}\right). \quad (3)$$

**Sample $\theta_{tk}$:** Due to the Markovian construction, it is necessary to consider both backward and forward information for the inference of $\theta_{tk}$. Starting from the last time point $t=T$, one has $n_{\cdot Tk} \sim \text{Pois}(\lambda_k\theta_{Tk})$, $\theta_{Tk} \sim \text{Gam}(\theta_{(T-1)k},1/c_T)$. As $\theta_{Tk}$ is not linked to a future factor score, it can be directly marginalized out, leading to $n_{\cdot Tk} \sim \text{NB}(\theta_{(T-1)k},p_{Tk})$, where $p_{Tk} := \frac{\lambda_k}{c_T+\lambda_k}$. The NB distribution can be further augmented with an auxiliary count random variable as $l_{Tk} \sim \text{CRT}(n_{\cdot Tk},\theta_{(T-1)k})$, $n_{\cdot Tk} \sim \text{NB}(\theta_{(T-1)k},p_{Tk})$. Following Lemma 2.1, the joint distribution of $l_{Tk}$ and $n_{\cdot Tk}$ is a bivariate count distribution that can be equivalently represented as $n_{\cdot Tk} \sim \sum_{t=1}^{l_{Tk}}\text{Log}(p_{Tk})$, $l_{Tk} \sim \text{Pois}(-\theta_{(T-1)k}\ln(1-p_{Tk}))$. Thus $l_{Tk}$ can be considered as the backward information from $T$ to $(T-1)$. Recursively, given $l_{(t+1)k}$, the backward information from $(t+1)$ to $t$, we then have:

$$l_{(t+1)k} \sim \text{Pois}(-\theta_{tk}\ln(1-p_{(t+1)k})),\ n_{\cdot tk} \sim \text{Pois}(\lambda_k\theta_{tk}).$$

The marginalization of $\theta_{tk}$ leads to

$$(n_{\cdot tk}+l_{(t+1)k}) \sim \text{NB}(\theta_{(t-1)k},p_{tk}),$$

where $p_{tk} := \frac{\lambda_k-\ln(1-p_{(t+1)k})}{c_t+\lambda_k-\ln(1-p_{(t+1)k})}$. Thus $l_{tk}$, the backward information to $(t-1)$, can be sampled as

$$(l_{tk}|-) \sim \text{CRT}(n_{\cdot tk}+l_{(t+1)k},\theta_{(t-1)k}). \quad (4)$$

With these information calculated *backwards* from $t=T$ to $t=1$, one can then sample $\theta_{tk}$ *forwards* from $t=0$ to $t=T$ as

$$(\theta_{tk}|-) \sim \text{Gam}(\theta_{(t-1)k}+n_{\cdot tk}+l_{(t+1)k},(1-p_{tk})/c_t). \quad (5)$$

where $n_{\cdot 0k}:=0$ and $\theta_{(-1)k}:=0.01$. This unique sampling procedure effectively solves the challenge of inferring the gamma shape parameters in a Markov chain.

**Sample $c_t$, $c$ and $\gamma_0$:** For $t=0,\ldots,T$, we sample $c_t$ as

$$(c_t|-) \sim \text{Gam}\left(e_0+\sum_k \theta_{(t-1)k},\frac{1}{f_0+\sum_k \theta_{tk}}\right). \quad (6)$$

We sample $c$ as

$$(c|-) \sim \text{Gam}\left(e_0+\gamma_0,\frac{1}{f_0+\sum_k \lambda_k}\right). \quad (7)$$

Since $n_{\cdot\cdot k} \sim \text{NB}\left(\frac{\gamma_0}{K},\frac{\sum_t \theta_{tk}}{c+\sum_t \theta_{tk}}\right)$, $\gamma_0$ can be sampled as:

$$(\ell_k|-) \sim \text{CRT}\left(n_{\cdot\cdot k},\frac{\gamma_0}{K}\right), \quad (8)$$
$$(\gamma_0|-) \sim \text{Gam}\left(e_0+\sum_k \ell_k,\frac{1}{f_0-\sum_k \frac{1}{K}\ln\left(1-\frac{\sum_t \theta_{tk}}{c+\sum_t \theta_{tk}}\right)}\right).$$

### 3.2 Modeling Binary Observations

To model binary data, a data augmentation technique is introduced here. Rather than following the usual approach to link a binary observation to a latent Gaussian random variable using the probit or logit links, a binary observation is linked to a latent count as

$$b = \mathbf{1}(n\geq 1),\ n\sim\text{Pois}(\lambda),$$

which is named in this paper as the Bernoulli-Poisson (BePo) link. We call the distribution of $b$ given $\lambda$ as the BePo distribution, with PMF $f_B(b|\lambda) = e^{-\lambda(1-b)}(1-e^{-\lambda})^b$, $b\in\{0,1\}$. The conditional posterior of the latent count $n$ is simply $(n|b,\lambda) \sim b\cdot\text{Pois}_+(\lambda)$, where $k\sim\text{Pois}_+(\lambda)$ is the truncated Poisson distribution with PMF $f_K(k) = \frac{1}{1-e^{-\lambda}}\frac{\lambda^k e^{-\lambda}}{k!}$, $k=1,2,\ldots$. Thus if $b=0$, then $n=0$ almost surely (a.s.), and if $b=1$, then $n$ is drawn from a truncated Poisson distribution. To simulate the truncated Poisson random variable $x\sim\text{Pois}_+(\lambda)$, we use rejection sampling: if $\lambda\geq 1$, we draw $x\sim\text{Pois}(\lambda)$ till $x\geq 1$; if $\lambda<1$, we draw $y\sim\text{Pois}(\lambda)$ and $u\sim\text{Unif}(0,1)$, and let $x=y+1$ if $u<1/(y+1)$. The acceptance rate is $1-e^{-\lambda}$ if $\lambda\geq 1$ and $\lambda^{-1}(1-e^{-\lambda})$ if $\lambda<1$. Thus the minimum acceptance rate is 63.2% (when $\lambda=1$).

With the BePo link to connect an observed dynamic binary matrix to a dynamic latent count matrix, we

Ayan Acharya, Joydeep Ghosh, Mingyuan Zhou

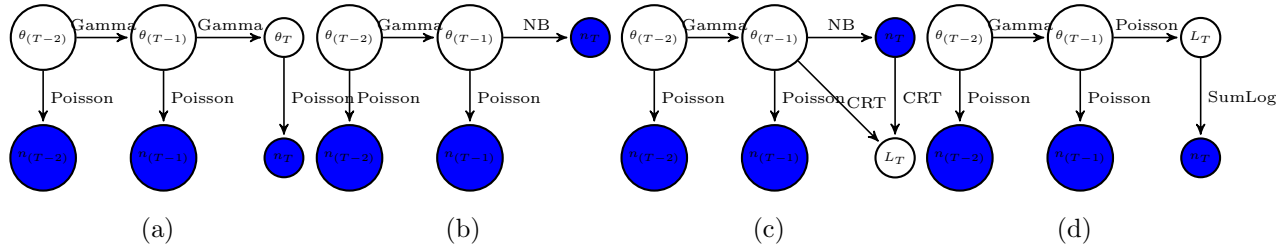

(a)    (b)    (c)    (d)

Figure 1: Illustration of Inference in GPAR

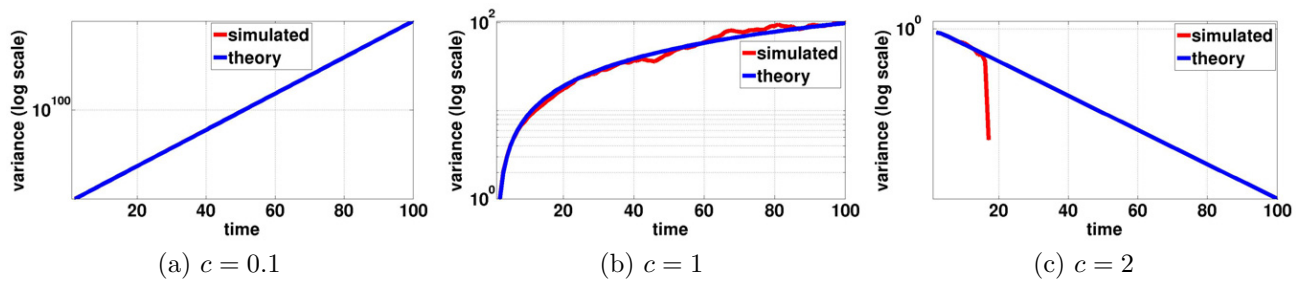

(a) $c = 0.1$    (b) $c = 1$    (c) $c = 2$

Figure 2: Plots of Variance

apply GP-DPFA to dynamic binary matrix factorization. The only additional step is to add the sampling of the latent counts as

$$(n_{vt}|b_{vt}, \mathbf{\Phi}, \mathbf{\Theta}, \mathbf{\Lambda}) \sim b_{vt}\text{Pois}_+(\sum_k \lambda_k \phi_{vk} \theta_{tk}). \quad (9)$$

A clear advantage of the BePo link over both the probit and logit links is that it is extremely efficient in handling sparse matrices: if an element of the binary matrix is zero, then the corresponding latent count under the BePo link is zero a.s., whereas further calculation is often required to sample the corresponding latent variable under both the probit and logit links.

### 3.3 Gamma Poisson Auto-Regressive Model

A special case of GP-DPFA with $K = V = 1$ is named as the gamma-Poisson auto-regressive model (GPAR). A precise description of this model is expressed as

$$n_t \sim \text{Pois}(\theta_t), \; \theta_t \sim \text{Gam}(\theta_{(t-1)}, 1/c), \; c \sim \text{Gam}(e_0, 1/f_0).$$

In Section 4, we evaluate the performance of GPAR and compare it with some interesting baselines. Analyzing the properties of GPAR is also interesting in its own right. The data augmentation used for inference in GP-DPFA can be better illustrated in the context of GPAR. Fig. 1(a) shows a segment of the GPAR model corresponding to time $t = T$, $t = (T - 1)$ and $t = (T - 2)$. Marginalization of $\theta_T$ leads to the model in Fig. 1(b) with $n_T$ being generated from $\theta_{(T-1)}$ using an NB distribution. Further, in Fig. 1(c), a latent CRT variable $L_T$ is augmented whose parameters are given by $\theta_{(T-1)}$ and $n_T$. Using the compound Poisson representation of the NB distribution, one can show that the model in Fig. 1(d) is equivalent to the model in Fig. 1(c) as far as the joint distribution of $L_T$ and $n_T$ is concerned. In the inference, once $L_T$ is sampled from $\text{CRT}(n_T, \theta_{(T-1)})$, sampling of $\theta_{(T-1)}$ becomes straightforward as the conditional distribution of $\theta_{(T-1)}$ given $n_{(T-1)}$ and $L_T$ is a gamma distribution.

The predictive performance of GPAR can be investigated using the laws of total expectation and total variance. Given observations up to time $T$, the expected latent rate at time $(t+T)$ is $\mathbb{E}[\theta_{T+t}] = c^{-t}\mathbb{E}[\theta_T]$, where $c$ can be inferred from training data and held fixed in prediction. Similarly, using the laws of total variance, the variance of the latent rate at time $(t + T)$ is

$$\text{Var}(\theta_{T+t}) = c^{-t}\left(\sum_{t'=1}^{t} c^{-t'}\right) \mathbb{E}(\theta_T). \quad (10)$$

Fig. 2 shows the variance for $c \in \{0.1, 1, 2\}$. The plot marked as "theory" displays the variances dictated by (10) and the one marked as "simulated" illustrates the sample variances of 10,000 $i.i.d.$ random samples simulated from the gamma Markov chain. Note that, as $t$ gets larger, the simulated plot deviates from the theoretical one in Fig. 2(c) due to numerical issues on a finite-precision machine. The inferred posterior of $c$ can be used to indicate the model's prediction of the trend of $\theta_t$. For example, if the posterior high density region of $c$ is above one, then $\theta_t$ would be predicted to have a clear downtrend.

## 4 EXPERIMENTS

In this section, experimental results are reported on a variety of synthetic and real world datasets and GP-DPFA is compared with relevant baselines. The synthetic and coal-mine disaster datasets provide a testbed of GPAR, a special case of GP-DPFA.

Table 2: Results on Synthetic Data

| Data | Measure | GPAR | SGCP | KS | LGCP10 | LGCP100 |
|---|---|---|---|---|---|---|
| SDS1 | MSE | **4.18**±0.04 | 4.20±0.03 | 6.65±0.10 | 5.96±0.07 | 5.44±0.12 |
|  | PMSE | **2.82**±0.09 | 3.20±0.11 | 5.43±0.19 | 6.92±0.20 | 4.28±0.37 |
| SDS2 | MSE | **27.12**±0.15 | 38.38±0.13 | 73.71±0.14 | 70.34±0.12 | 43.51±0.07 |
|  | PMSE | **10.14**±0.31 | 12.01±0.22 | 13.49±0.38 | 14.73±0.22 | 12.52±0.43 |
| SDS3 | MSE | **10.94**±0.07 | 11.41±0.05 | 30.56±0.16 | 90.76±0.14 | 10.79±0.14 |
|  | PMSE | **5.81**±0.16 | 7.19±0.12 | 25.17±0.41 | 28.72±0.18 | 20.08±0.52 |

Table 3: Results on Real-world Text Data

| Data | Model | MP | MR | PP |
|---|---|---|---|---|
| STU | GP-DPFA | **0.2230**±0.0009 | 0.1976±0.0004 | **0.1891**±0.0028 |
|  | DRFM | 0.2171±0.0025 | **0.1978**±0.0014 | 0.1773±0.0104 |
|  | Baseline | 0.1018±0.0216 | 0.1329±0.0173 | 0.0612±0.0328 |
| Conf. | GP-DPFA | 0.3020±0.0004 | **0.2681**±0.0003 | **0.2412**±0.0004 |
|  | DRFM | **0.3023**±0.0005 | 0.2566±0.0006 | 0.2410±0.0006 |
|  | Baseline | 0.1241±0.0194 | 0.1107±0.0131 | 0.1014±0.0370 |

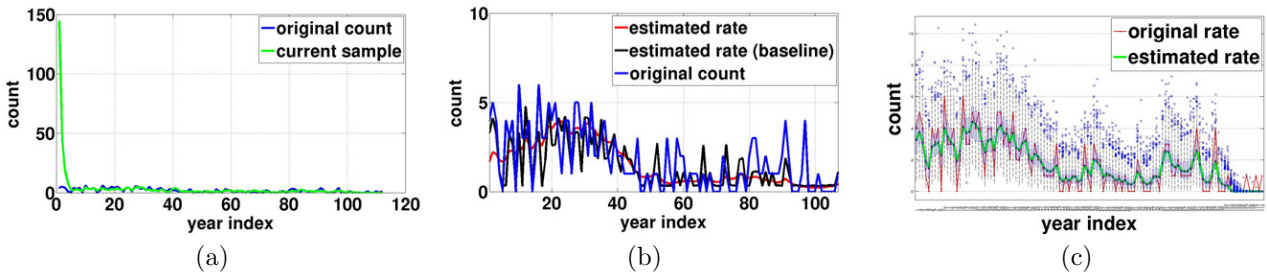

(a)   (b)   (c)

Figure 3: (a) estimate of the underlying rate after first iteration, (b) estimate of the rate after 3000 iterations (with the first 2000 iterations used as burnin), (c) estimate of the rate with uncertainty after 3000 iterations.

### 4.1 Results with Synthetic Datasets

As in [35], three one-dimensional data sets are used with the following rate functions:

• A sum of an exponential and a Gaussian bump (SDS1): $\theta(t) = 2\exp(-t/15) + \exp(-((t-25)/10)^2)$ on the interval $t = [0:1:50]$.

• A sinusoid with increasing frequency (SDS2): $\theta(t) = 5\sin(t^2) + 6$ on $t = [0:0.2:5]$.

• $\theta$ is the piecewise linear function on the interval $t = [0:1:100]$ and is given by: $\theta(t) = (2 + t/30)$ if $0 \leq t \leq 30$, $\theta(t) = (3 - (t-30)/10)$ if $31 \leq t \leq 50$, $\theta(t) = (1 + 1.5 * (t-50)/25)$ if $51 \leq t \leq 75$ and $\theta(t) = (2.5 + 0.5 * (t-75)/25)$ if $76 \leq t \leq 100$ (SDS3).

GPAR is compared with the sigmoidal Gaussian Cox process (SGCP) [35], log-Gaussian Cox process (LGCP) [36], and the classical kernel smoothing (KS) [37]. These methods are considered as state-of-the-art in various scenarios involving modeling of count time series. Edge-corrected kernel smoothing is performed using a quartic kernel and a mean-square minimization technique is used for bandwidth selection. The squared-exponential kernel is used for both the SGCP and LGCP. Since the LGCP works with discretization, experiments are performed with 10, 25 and 100 bins. The rate functions provide ground truth and cumulative mean squared error (MSE) between the ground truth and the estimated rate are measured for all the models. Additionally, for each of the above series, the last five observations are withheld and MSE is measured between the true rate and the estimated rate over these withheld observations. The results are displayed in Table 2. "PMSE" stands for MSE in prediction for the last five years of data. The best results are presented in bold.

### 4.2 Results with Real World Datasets

**Coalmine Disaster Dataset:** The British coal mine disaster dataset [35] records the number of coalmine accidents arranged according to year from 1851 to 1962. To illustrate the robustness of the inference framework, the underlying rate is initialized to a large value 1000. Fig. 1(a) shows the estimated rate and the sampled value of the underlying rate after the 1$^{st}$ iteration. Fig. 1(b) shows the estimation of the underlying rate along with a "baseline" GP-DPFA model that does not use any temporal correlation. A box plot of the sampled rate is presented in Fig. 1(c) showing that the alogorithm converges to a good estimate even with such a poor initialization. For these plots, 3000 iterations are used and the last 1000 samples are collected.

**State-of-the-Union Dataset (STU):** The STU dataset contains the transcripts of 225 US State of the Union addresses, from 1790 to 2014. Each transcript corresponding to each year is considered as one document. After removing stop words and terms that occur fewer than 7 times in one document or less than 20 times overall, there are 2375 unique words.

**Conference Abstract Dataset (Conf.):** The Conf. dataset consists of the abstracts of the papers appearing on DBLP for the second author of this paper from 2000 to 2013. For every year, a count vector of dimension $V = 1771$ is maintained where the counts are the occurrences of the words appearing in all documents from the given year, chosen after standard preprocessing like stemming and stop-words removal.

Table 3 displays the results from both STU and Conf. datasets. 20% of the words are held-out for each of the first 224 years in the STU data and 10% of the words are held-out for each of the first 13 years in the Conf. data, when training three different models: i) GP-DPFA, ii) DRFM [26], and iii) a baseline model which is a simplified version of GP-DPFA that does not use temporal correlation for the latent rates. Additionally, all the data from the last year for both of these datasets are held-out. The underlying prediction problem is concerned with estimating the held-out words. For the prediction corresponding to each year, the words are ranked according to the estimated count and then two quantities are calculated: i) precision@top-$M$ which is given by the fraction of the top-$M$ words, predicted by the model, that matches the true ranking of the words; and ii) recall@top-$M$ which is given by the fraction of words from the held-out set that appear in the top$-M$ ranking. In the experiments reported, $M = 50$ is used. For the last year for which entire data is held-out, calculation of recall@top$-M$ is irrelevant. In Table 3, the column MP and MR signify mean precision and mean recall respectively over all the years that appear in the training set. The column PP signifies the predictive precision for the final year, for which the entire dataset is held out. Such measure is also adopted for the recommendation system in [22] and is perhaps the only reasonable measure when the likelihoods between two different models like GP-DPFA and DRFM are not comparable. GP-DPFA almost always outperforms DRFM and both of these dynamic models convincingly beat the baseline model.

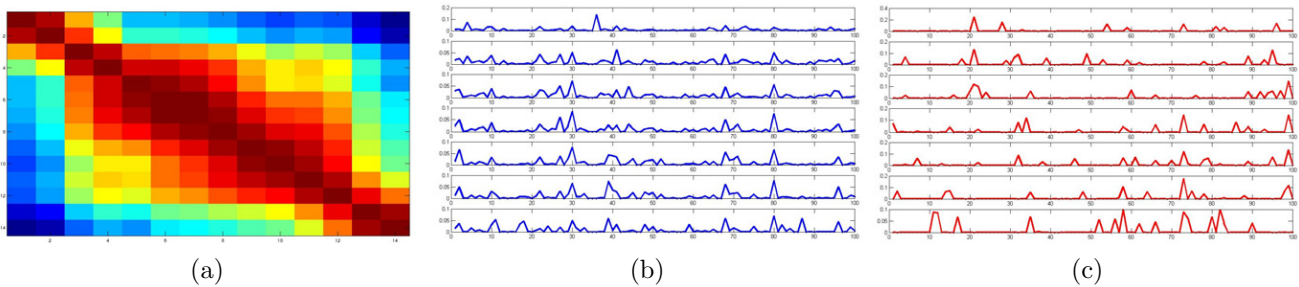

(a) (b) (c)

Figure 4: (a) correlation across topics over time, (b) latent factors dominant over time for GP-DPFA, (c) latent factors dominant over time for the baseline.

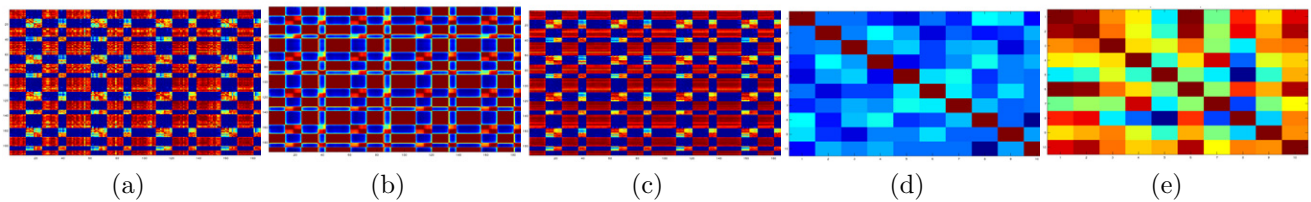

(a) (b) (c) (d) (e)

Figure 5: (a) correlation of the observed data across time, (b) correlation discovered in the latent space, (c) correlation between the observation and latent counts, (d) correlation between the ten most prominent latent factors for GP-DPFA, (e) correlation between the ten most prominent latent factors in the baseline model.

For the Conf. dataset, Fig. 4(a) shows the correlation discovered in the latent space over time, and Figs. 4 (b) and (c) show the normalized strengths of the latent factors (i.e. $\lambda_k \theta_{tk}/\sum_k \lambda_k \theta_{tk}$) discovered by GP-DPFA and the baseline model, respectively. One can clearly see that the assignments to latent factors are strongly correlated with time for GP-DPFA but the baseline model tends to choose different latent factors for different years. In the experiments, $K = 100$ is used and GP-DPFA infers that only a small subset of the 100 topics need to be active, implying an automatic model selection. The number of active latent topics is found to be around 14 on average. Examining some of the topics provides even more insight about the data. For example, the top words of a topic that has large weights across all years include "network", "graph-partition", "algorithm", "cluster" and "outlier", whereas the top words of a topic that is dominant over a certain period of time include "Bregman", "projection", "clustering" and "ensemble", revealing the author's publication trend.

**Music Dataset**: Four different polyphonic music sequences of piano are used for experiments with GP-DPFA. Each of these datasets is a collection of binary strings indicating which of the keys are "on" at each time [38, 39]. "Nottingham" is a collection of 1200 folk tunes, "Piano.midi" is a classical piano MIDI archive, "MuseData" is an electronic library of orchestral and piano classical music, and "JSB chorales" refers to the entire corpus of 382 four-part harmonized chorales by J. S. Bach. The polyphony (number of simultaneous notes) varies from 0 to 15 and the average polyphony is 3.9. We use an input of 88 binary units that span the whole range of piano from A0 to C8. In Fig. 5, re-sults are displayed for one of the 1200 tunes from Nottingham data. Fig. 5(a) shows the correlation of the binary strings across time. Interestingly, a similar but more prominent correlation structure is discovered in the latent factor scores (*i.e.* across $(\lambda_k \theta_{tk})_{k=1}^{K}$'s), displayed in Fig. 5(b). Additionally, the correlations between the original data and the estimated latent counts are presented in Fig. 5(c). One can see that this correlation plot perfectly imitates the correlation between the original data, implying that the original data are faithfully reconstructed using GP-DPFA. Also, in Fig. 5(d) we display the correlation between the top ten $\phi_k$'s (ranked according to the magnitudes of the $\lambda_k$'s) discovered by GP-DPFA. We compare this plot with Fig. 5(e), which shows the correlation between the top ten $\phi_k$'s discovered by the non-dynamic baseline model. One can clearly see that GP-DPFA discovers comparatively less correlated latent factors.

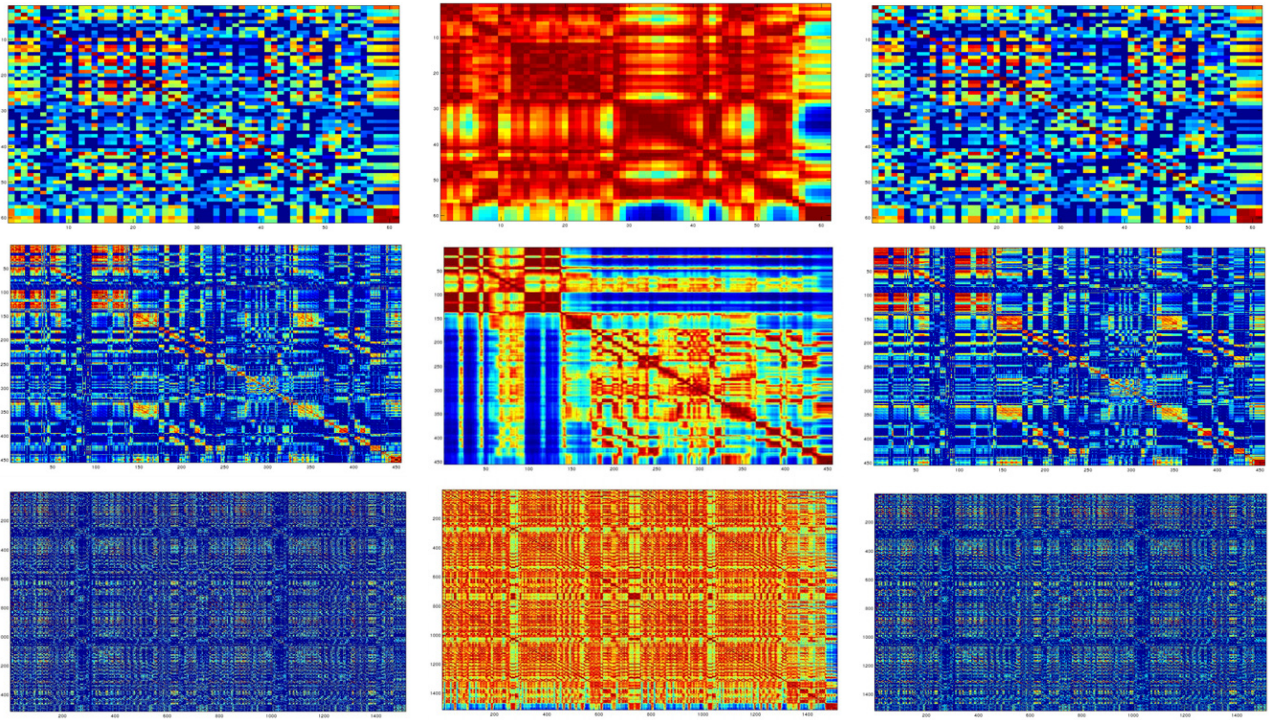

Figure 6: Top Row: Correlation plots for JSB chorales, Middle Row: Correlation plots for Piano.midi, Bottom Row: Correlation plots for Musedata. In each row, figures from left to right are plots that are analogous to Figs. 5 (a)-(c).

The top, middle and bottom rows in Fig. 6 illustrate the correlation plots for one of the tunes in the JSB chorales dataset, the Piano.midi data and the MuseData, respectively. The left-most plot in each of the rows shows the correlation of the observed data. The plots in the middle illustrate the correlation discovered in the latent space and the plots in the last column shows the the correlation between the observed data and estimated latent counts. It is shown that even when the correlation structure is not clear in the original data, very clear correlation structure is discovered in the latent space, without sacrificing the data reconstruction quality.

## 5 CONCLUSIONS

This paper introduces gamma process dynamic Poisson factor analysis to model multivariate count and binary vectors that evolve over time. The constructed gamma Markov chain is a unique contribution of the paper. Efficient inference techniques and superior empirical performance on both synthetic and real world datasets make the approach a promising candidate for modeling more sophisticated count time-series data; for example, time-evolving matrices and tensors that appear quite frequently in social network analysis, text mining and recommendation systems.